\documentclass[10pt,twocolumn,letterpaper]{article}

\usepackage{cvpr}
\usepackage{times}
\usepackage{epsfig}
\usepackage{graphicx}
\usepackage{amsmath}
\usepackage{amssymb}
\usepackage{xcolor}
\usepackage{algorithm} % Boxes/formatting around algorithms
\usepackage{algpseudocode} % Algorithms
\usepackage{paralist}
\usepackage{enumitem}
\usepackage{subcaption}

\DeclareMathOperator*{\argmin}{arg\,min}

\usepackage[pagebackref=true,breaklinks=true,letterpaper=true,colorlinks,bookmarks=false]{hyperref}

\cvprfinalcopy % *** Uncomment this line for the final submission

\def\cvprPaperID{****} % *** Enter the CVPR Paper ID here

\ifcvprfinal\fi
\begin{document}

%Highlight boxes
\newcommand{\hly}{\colorbox{yellow}}
\newcommand{\hlg}{\colorbox{green}}
\newcommand{\hlp}{\colorbox{pink}}

\title{Fast Underwater Image Enhancement for Improved Visual Perception}

\author{Md Jahidul Islam$^1$, Youya Xia$^2$ and Junaed Sattar$^3$ \\
{\tt\small \{$^1$islam034,$^2$xiaxx244,$^3$junaed\}@umn.edu} \\
{\small Interactive Robotics and Vision Laboratory, Department of Computer Science and Engineering} \\ 
{\small Minnesota Robotics Institute, University of Minnesota, Twin Cities, MN, USA }
}

\maketitle
%\thispagestyle{empty}

%%%%%%%%% source files %%%%%%%%

\begin{abstract}
In this paper, we present a conditional generative adversarial network-based model for real-time underwater image enhancement. To supervise the adversarial training, we formulate an objective function that evaluates the perceptual image quality based on its global content, color, local texture, and style information. We also present EUVP, a large-scale dataset of a paired and an unpaired collection of underwater images (of `poor' and `good' quality) that are captured using seven different cameras over various visibility conditions during oceanic explorations and human-robot collaborative experiments. In addition, we perform several qualitative and quantitative evaluations which suggest that the proposed model can learn to enhance underwater image quality from both paired and unpaired training. More importantly, the enhanced images provide improved performances of standard models for underwater object detection, human pose estimation, and saliency prediction. These results validate that it is suitable for real-time preprocessing in the autonomy pipeline by visually-guided underwater robots. The model and associated training pipelines are available at \url{https://github.com/xahidbuffon/funie-gan}. 
\end{abstract}

\section{Introduction}
% why do we need this: challenges and scope
Visually-guided AUVs (Autonomous Underwater Vehicles) and ROVs (Remotely Operated Vehicles) are widely used in important applications such as the monitoring of marine species migration and coral reefs~\cite{shkurti2012multi}, inspection of submarine cables and wreckage~\cite{bingham2010robotic}, underwater scene analysis, seabed mapping, human-robot collaboration~\cite{islam2018understanding}, and more. One major operational challenge for these underwater robots is that despite using high-end cameras, visual sensing is often greatly affected by poor visibility, light refraction, absorption, and scattering~\cite{lu2013underwater,zhang2017underwater,islam2018understanding}. These optical artifacts trigger non-linear distortions in the captured images, which severely affect the performance of vision-based tasks such as tracking, detection and classification, segmentation, and visual servoing. Fast and accurate image enhancement techniques can alleviate these problems by restoring the perceptual and statistical qualities~\cite{fabbri2018enhancing,zhang2017underwater} of the distorted images in real-time.           

\begin{figure}[t]
    \centering
    \begin{subfigure}{0.48\textwidth}
        \includegraphics[width=\linewidth]{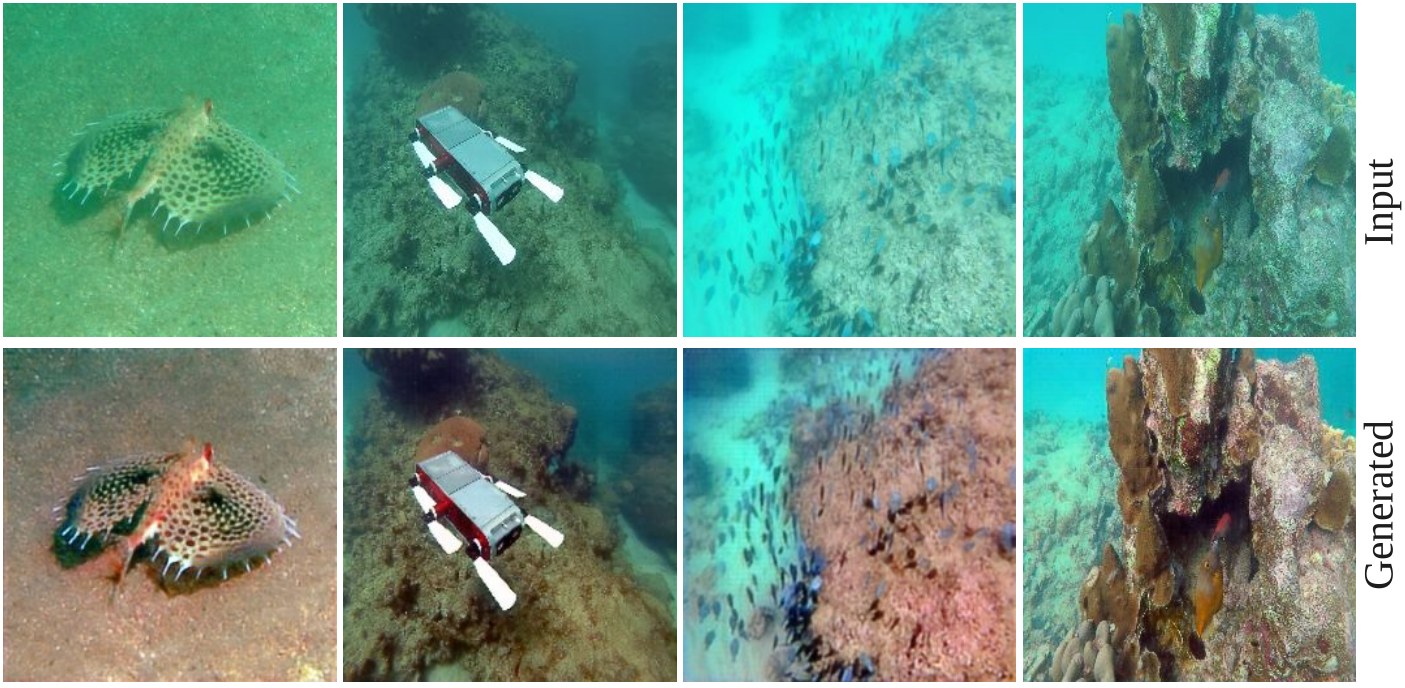}% 
        \vspace{-1mm}
        \caption{Perceptual enhancement of underwater images.}%
    \end{subfigure}
    
    \vspace{1mm}
    \begin{subfigure}{0.48\textwidth}
        \includegraphics[width=\linewidth]{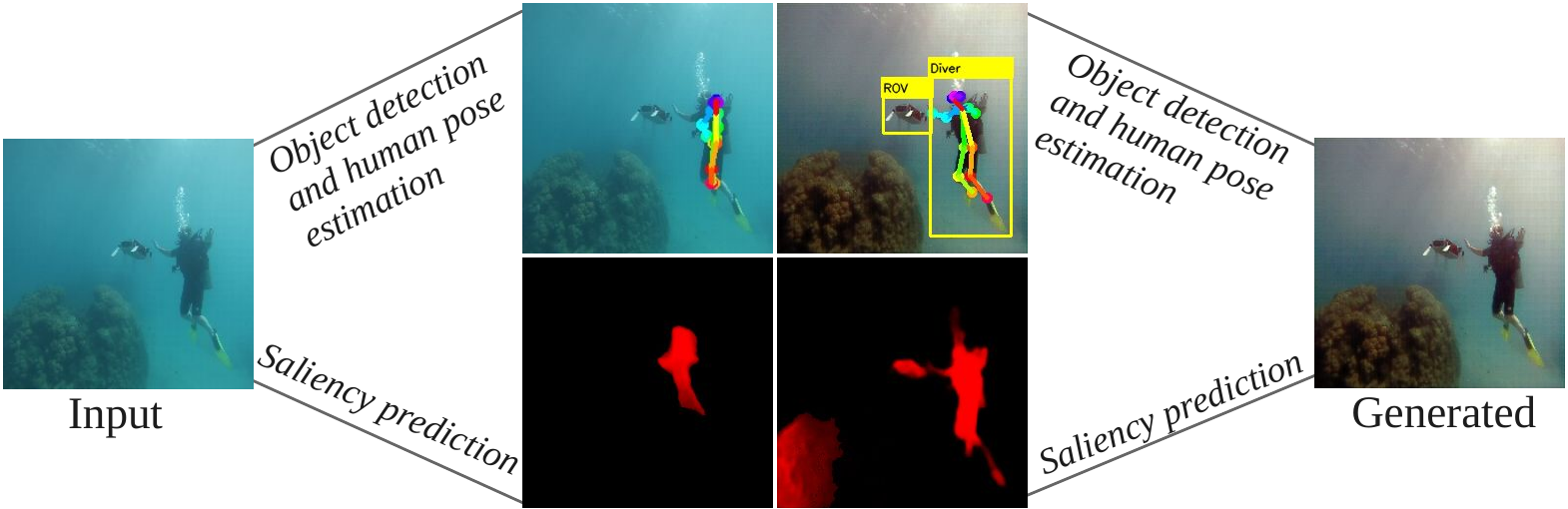}% 
        \vspace{-1mm}
        \caption{Improved performance for underwater object detection~\cite{islam2018towards}, human body-pose estimation~\cite{cao2017realtime}, and saliency prediction~\cite{wang2018salient}.}
    \end{subfigure}
    
    \caption{Demonstration of underwater image enhancement using our proposed model and its practical feasibility.}
    \label{fig:1}
\end{figure}

% physics-based to learning-based
As light propagation differs underwater (than in the atmosphere), a unique set of non-linear image distortions occur which are propelled by a variety of factors. 
For instance, underwater images tend to have a dominating green or blue hue~\cite{fabbri2018enhancing} because red wavelengths get absorbed in deep water (as light travels further). Such wavelength dependant attenuation~\cite{akkaynak2018revised}, scattering, and other optical properties of the waterbodies cause irregular non-linear distortions~\cite{guo2019underwater,zhang2017underwater} which result in low-contrast, often blurred, and color-degraded images. Some of these aspects can be modeled and well estimated by physics-based solutions, particularly for dehazing and color correction~\cite{bryson2016true,berman2018underwater}. However, information such as the scene depth and optical water-quality measures are not always available in many robotic applications. Besides, these models are often computationally too demanding for real-time deployments.

% learning-based to our contributions
A practical alternative is to approximate the underlying solution by learning-based methods, which demonstrated remarkable success in recent years. Several models based on deep Convolutional Neural Networks (CNNs) and Generative Adversarial Networks (GANs) provide state-of-the-art performance~\cite{ignatov2017dslr,chen2018deep,isola2017image,zhu2017unpaired} in learning to enhance perceptual image quality from a large collection of paired or unpaired data. 
For underwater imagery, in particular, a number of GAN-based models~\cite{fabbri2018enhancing,yu2018underwater} and CNN-based residual models~\cite{liu2019underwater} report inspiring progress for automatic color enhancement, dehazing, and contrast adjustment. However, there is significant room for improvement as learning perceptual enhancement for underwater imagery is a more challenging ill-posed problem (than terrestrial imagery).    
Additionally, due to the high costs and difficulties associated with acquiring large-scale underwater data, most learning-based models use small-scale and often only synthetically generated images that fail to capture a wide range of natural variability. 
Moreover, designing robust yet efficient image enhancement models and investigating their applicability for improving real-time underwater visual perception have not been explored in the literature in depth. 

%such as color/contrast adjustment~\cite{isola2017image,cheng2015deep}, colorization~\cite{zhang2016colorful}, dehazing~\cite{cai2016dehazenet}, super-resolution~\cite{dong2014learning}, etc. 

%% contributions
We attempt to address these challenges by designing a fast underwater image enhancement model and analyzing its feasibility for real-time applications. We formulate the problem as an image-to-image translation problem by assuming there exists a non-linear mapping between the distorted (input) and enhanced (output) images.  
Then, we design a conditional GAN-based model to learn this mapping by adversarial training on a large-scale dataset named EUVP (Enhancement of Underwater Visual Perception). 
%see enumitem package to use custom enumerate [topsep=1pt,itemsep=0.5pt,parsep=1pt]
From the perspective of its design, implementation, and experimental validation, we make the following contributions in this paper: 
\begin{compactenum}[(a)]
\vspace{1mm}
\item We present a fully-convolutional conditional GAN-based model for real-time underwater image enhancement, which we refer to as FUnIE-GAN. We formulate a multi-modal objective function to train the model by evaluating the perceptual quality of an image based on its global content, color, local texture, and style information.    

\vspace{1mm}
\item Additionally, we present the EUVP dataset, a paired and an unpaired collection of $20$K underwater images (of poor and good quality) that can be used for \textit{one-way} and \textit{two-way} adversarial training~\cite{chen2018deep,zhu2017unpaired}. The dataset is available at {\tt\url{http://irvlab.cs.umn.edu/resources/euvp-dataset}}.

\vspace{1mm}
\item Furthermore, we present qualitative and quantitative performance evaluations compared to state-of-the-art models. The results suggest that FUnIE-GAN can learn to enhance perceptual image quality from both paired and unpaired training. More importantly, the enhanced images significantly boost the performance of several underwater visual perception tasks such as object detection, human pose estimation, and saliency prediction; a few sample demonstrations are highlighted in Fig.~\ref{fig:1}.
\end{compactenum}
\vspace{1mm}

In addition to presenting the conceptual model of FUnIE-GAN, we analyze important design choices and relevant practicalities for its efficient implementation. We also conduct a user study and a thorough feasibility analysis to validate its effectiveness for improving the real-time perception performance of visually-guided underwater robots. 

\section{Related Work}
%We now present a brief overview on the existing literature that are relevant to our problem of interest.

\subsection{Automatic Image Enhancement}
Automatic image enhancement is a well-studied problem in the domains of computer vision, robotics, and signal processing. Classical approaches use hand-crafted filters to enforce local color constancy and improve contrast/lightness rendition~\cite{rahman2004retinex}. Additionally, prior knowledge or statistical assumptions about a scene (\eg, haze-lines, dark channel prior~\cite{berman2018underwater}, etc.) are often utilized for global enhancements such as image deblurring,  dehazing~\cite{he2010single}, etc. Over the last decade, single image enhancement has made remarkable progress due to the advent of deep learning and the availability of large-scale datasets. The contemporary deep CNN-based models provide state-of-the-art performance for problems such as image colorization~\cite{zhang2016colorful}, color/contrast adjustment~\cite{cheng2015deep}, dehazing~\cite{cai2016dehazenet}, etc. These models learn a sequence of non-linear filters from paired training data, which provide much better performance compared to using hand-crafted filters.

Moreover, the GAN-based models~\cite{goodfellow2014generative} have shown great success for style-transfer and image-to-image translation problems~\cite{isola2017image}. They employ a two-player min-max game where the `generator' tries to fool the `discriminator' by generating \textit{fake} images that appear to be sampled from the \textit{real} distribution. Simultaneously, the discriminator tries to get better at discarding fake images and eventually (in equilibrium) the generator learns to model the underlying distribution. Although such adversarial training can be unstable, several tricks and choices of loss functions are proposed in the literature to mitigate that. For instance, Wasserstein GAN~\cite{arjovsky2017wasserstein} improves the training stability by using the earth-mover distance to measure the distance between the data distribution and the model distribution. Energy-based GANs~\cite{zhao2016energy} also improve training stability by modeling the discriminator as an energy function, whereas the Least-Squared GAN~\cite{mao2017least} addresses the vanishing gradients problem by adopting a least-square loss function for the discriminator. On the other hand, conditional GANs~\cite{mirza2014conditional} allow constraining the generator to produce samples that follow a pattern or belong to a specific class, which is particularly useful to learn a pixel-to-pixel (Pix2Pix) mapping~\cite{isola2017image} between an arbitrary input domain (\eg, distorted images) and the desired output domain (\eg, enhanced images).

\begin{figure*}[t]
    \centering
    \begin{subfigure}{0.62\textwidth}
        \includegraphics[width=\linewidth]{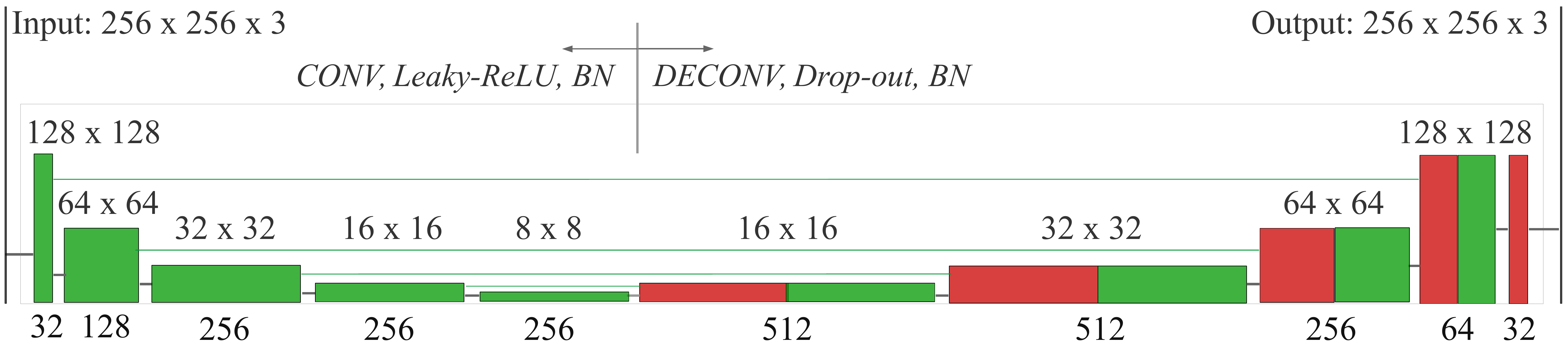}
        \caption{Generator: five encoder-decoder pairs with mirrored skip-connections (inspired by the success of U-Net~\cite{ronneberger2015u}; however, it is a much simpler model).}
        \label{fig:model_a}
    \end{subfigure}
    ~
    \begin{subfigure}{0.36\textwidth}
        \includegraphics[width=\linewidth]{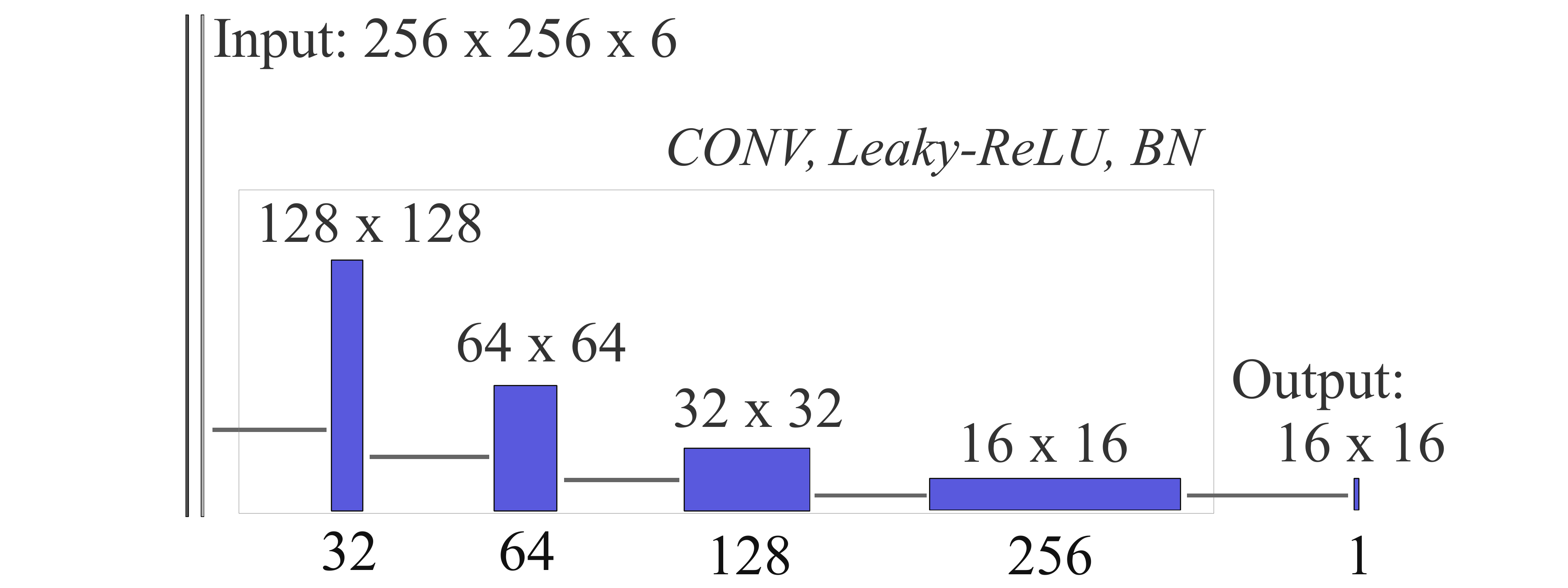}
        \caption{Discriminator: a Markovian PatchGAN~\cite{isola2017image} with four layers and a patch-size of $16$$\times$$16$.}
        \label{fig:model_b}
    \end{subfigure}
    \caption{Network architecture of the proposed model: FUnIE-GAN.}
    \label{fig:model}
\end{figure*}

A major limitation of the above-mentioned models is that they require paired training data, which may not be available or can be difficult to acquire for many practical applications. The two-way GANs (\eg, CycleGAN~\cite{zhu2017unpaired}, DualGAN~\cite{yi2017dualgan}, etc.) solve this problem by using a `cycle-consistency loss' that allows learning the mutual mappings between two domains from unpaired data. Such models have been effectively used for unpaired learning of perceptual image enhancement~\cite{chen2018deep} as well. Furthermore, Ignatov \etal~\cite{ignatov2017dslr} showed that additional loss-terms for preserving the high-level feature-based content improve the quality of image enhancement using GANs.

\subsection{Improving Underwater Visual Perception}
Traditional physics-based methods use the atmospheric dehazing model to estimate the \textit{transmission} and \textit{ambient} light in a scene to recover true pixel intensities~\cite{cho2018model,bryson2016true}. 
Another class of methods design a series of bilateral and trilateral filters to reduce noise and improve global contrast~\cite{lu2013underwater,zhang2017underwater}. In recent work, Akkaynak\etal~\cite{akkaynak2018revised} proposed a revised imaging model that accounts for the unique distortions pertaining to underwater light propagation; this contributes to a more accurate color reconstruction and overall a better approximation to the ill-posed underwater image enhancement problem. Nevertheless, these methods require scene depth (or multiple images) and optical waterbody measurements as prior.

On the other hand, several single image enhancement models based on deep adversarial~\cite{fabbri2018enhancing,yu2018underwater,li2018watergan} and residual learning~\cite{liu2019underwater} have reported inspiring results of late. However, they typically use only synthetically distorted images for paired training, which often limit their generalization performance. The extent of large-scale unpaired training on naturally distorted underwater images have not been explored in the literature. Moreover, most existing models fail to ensure fast inference on single-board robotic platforms, which limits their applicability for improving real-time visual perception. We attempt to address these aspects in this paper.

%For instance, Li \etal~\cite{li2018watergan} generate synthetic underwater images by aligning them with in-air RGBD images. On the other hand, Fabbri \etal~\cite{fabbri2018enhancing} perform style-transfer using CycleGAN to generate distorted underwater images for preparing paired training instances.

\section{Proposed Model and Dataset}
%The following sections present the network architecture and specification of the proposed GAN-based model for fast underwater image enhancement, which we refer to as FUnIE-GAN. In addition, we provide detailed information about the proposed EUVP dataset. 

\subsection{FUnIE-GAN Architecture}
Given a source domain $X$ (of distorted images) and desired domain $Y$ (of enhanced images), our goal is to learn a mapping $G: X \rightarrow Y$ in order to perform automatic image enhancement. We adopt a conditional GAN-based model where the generator tries to learn this mapping by evolving with an adversarial discriminator through an iterative min-max game. %A simplified sketch of the network architecture of FUnIE-GAN is presented in Fig.~\ref{fig:model}.
As illustrated in Fig.~\ref{fig:model}, we design a generator network by following the principles of U-Net~\cite{ronneberger2015u}. 
It is an encoder-decoder network ($e_1$-$e_5$,$d_1$-$d_5$) with connections between the mirrored layers, \ie, between ($e_1$, $d_5$), ($e_2$, $d_4$), ($e_3$, $d_2$), and ($e_4$, $d_4$). Specifically, the outputs of each encoders are concatenated to the respective mirrored decoders. This idea of \textit{skip-connections} in the generator network is shown to be very effective~\cite{isola2017image,chen2018deep,fabbri2018enhancing} 
for image-to-image translation and image quality enhancement problems. In FUnIE-GAN, however, we employ a much simpler model with fewer parameters in order to achieve fast inference. The input to the network is set to $256\times256\times3$ and the encoder ($e_1$-$e_5$) learns only $256$ feature-maps of size $8\times8$. The decoder ($d_1$-$d_5$) utilizes these feature-maps and inputs from the skip-connections to learn to generate a $256\times256\times3$ (enhanced) image as output. The network is fully-convolutional as no fully-connected layers are used. Additionally, 2D convolutions with $4\times4$ filters are applied at each layer, which is then followed by a Leaky-ReLU non-linearity~\cite{maas2013rectifier} and Batch Normalization (BN)~\cite{ioffe2015batch}. The feature-map sizes in each layer and other model parameters are annotated in Fig.~\ref{fig:model_a}.

For the discriminator, we employ a Markovian PatchGAN~\cite{isola2017image} architecture that assumes the independence of pixels beyond the patch-size, \ie, only discriminates based on the patch-level information. This assumption is important to effectively capture high-frequency features such as local texture and style~\cite{yi2017dualgan}. In addition, this configuration is computationally more efficient as it requires fewer parameters compared to discriminating globally at the image level. As shown in Fig.~\ref{fig:model_b}, four convolutional layers are used to transform a $256\times256\times6$ input (real and generated image) to a $16\times16\times1$ output that represents the averaged \textit{validity} responses of the discriminator. At each layer, $3\times3$ convolutional filters are used with a stride of $2$; then the non-linearity and BN are applied the same way as the generator. %Traditionally, PatchGANs use $70\times70$ patches for $256\times256$ images (\eg, in Pix2Pix~\cite{isola2017image}, DualGAN~\cite{yi2017dualgan}, etc.). However, we use a patch-size of only $16$$\times$$16$ in FUnIE-GAN.                         

%%%%%%%%%%%%%%%%%%%%%%%%%%%%%%%%%%%%%%%%%%%%%%%%%%%%%
\subsection{Objective Function Formulation}\label{obj_fun}
A standard conditional GAN-based model learns a mapping $G:\{X, Z\} \rightarrow Y$, where $X$ ($Y$) represents the source (desired) domain, and $Z$ denotes random noise. The conditional adversarial loss function~\cite{mirza2014conditional} is expressed as:   
\begin{equation} %\label{eq:cgan}
\centering
  \begin{aligned}
	\mathcal{L}_{cGAN}(G,D) &= \mathbb{E}_{X,Y} \big[\log D(Y)\big]  \\
	 &+ \mathbb{E}_{X,Y} \big[\log (1-D(X, G(X, Z)))\big]
  \end{aligned}
\end{equation}
Here, the generator $G$ tries to minimize $\mathcal{L}_{cGAN}$ while the discriminator $D$ tries to maximize it. In FUnIE-GAN, we associate three additional aspects, \ie, global similarity, image content, and local texture and style information in the objective to quantify perceptual image quality.    

\begin{itemize}
    \item \textbf{Global similarity}: existing methods have shown that adding an $L_1$ ($L_2$) loss to the objective function enables $G$ to learn to sample from a globally similar space in an $L_1$ ($L_2$) sense~\cite{isola2017image,yu2018underwater}. Since the $L_1$ loss is less prone to introduce blurring, we add the following loss term in the objective:     
        \begin{equation} \label{eq:l1}
        \centering
        	\mathcal{L}_{1}(G) = \mathbb{E}_{X,Y,Z} \big[\big|\big|Y-G(X, Z)\big|\big|_1\big]
        \end{equation}

    \item \textbf{Image content}: we add a \textit{content loss} term in the objective in order to encourage $G$ to generate enhanced image that has similar content (\ie, feature representation) as the target (\ie, real) image. Being inspired by~\cite{johnson2016perceptual,ignatov2017dslr}, we define the image content function $\Phi(\cdot)$ as the high-level features extracted by the {\tt block5\_conv2} layer of a pre-trained VGG-19 network. Then, we formulate the content loss as follows:     
            \begin{equation} \label{eq:lcontent}
           \centering
        	\mathcal{L}_{con}(G) = \mathbb{E}_{X,Y,Z} \big[\big|\big|\Phi (Y)-\Phi (G(X, Z)) \big|\big|_2\big]
        \end{equation}

   \item \textbf{Local texture and style}: as mentioned, Markovian PatchGANs are effective in capturing high-frequency information pertaining to the local texture and style~\cite{isola2017image}. Hence, we rely on $D$ to enforce the local texture and style consistency in adversarial fashion.
\end{itemize}

\subsubsection{Paired Training}
For paired training, we formulate an objective function that guides $G$ to learn to improve the perceptual image quality 
so that the generated image is close to the respective ground truth %$y \in Y$ 
in terms of its global appearance and high-level feature representation. On the other hand, $D$ will discard a generated image that has locally inconsistent texture and style. %Hence, it'll force the $G$ to improve the high-frequency features as well.         
Specifically, we use the following objective function for paired training: 
\begin{equation*} %\label{eq:pix2pix_final}
\centering
	G^* = \argmin\limits_{G}\max\limits_{D} \mathcal{L}_{cGAN}(G,D)+\lambda_1 \mathcal{L}_{1}(G)+\lambda_c \mathcal{L}_{con}(G) 
\end{equation*}
Here, $\lambda_1=0.7$ and $\lambda_c=0.3$ are scaling factors that we empirically tuned as hyper-parameters.

\subsubsection{Unpaired Training}
For unpaired training, we do not enforce the global similarity and content loss constraints as the pairwise ground truth is not available. Instead, the objective is to learn both the forward mapping $G_{F}:\{X, Z\} \rightarrow Y$ and the reconstruction $G_{R}:\{Y, Z\} \rightarrow X$ simultaneously by maintaining cycle-consistency. As suggested by Zhu \etal~\cite{zhu2017unpaired}, we formulate the cycle-consistency loss as follows:   
\begin{equation} \label{eq:cycle}
\centering
  \begin{aligned}
	\mathcal{L}_{cyc}(G_{F}, G_{R}) &= \mathbb{E}_{X,Y,Z} \big[\big|\big|X-G_{R}(G_{F}(X, Z))\big|\big|_1\big]  \\
	 &+ \text{ } \mathbb{E}_{X,Y,Z} \big[\big|\big|Y-G_{F}(G_{R}(Y, Z))\big|\big|_1\big]
  \end{aligned}
\end{equation}
Therefore, our objective for the unpaired training is: 
\begin{equation*} %\label{eq:pix2pix_final}
\centering
   \begin{aligned}
	G_{F}^*, G_{R}^* = &\argmin\limits_{G_{F}, G_{R}}\max\limits_{D_Y,D_X} \mathcal{L}_{cGAN}(G_{F}, D_Y) \text{ } \\ &+ \text{ } \mathcal{L}_{cGAN}(G_{R}, D_X) \text{ } + \lambda_{cyc} \mathcal{L}_{cyc}(G_{F}, G_{R})
   \end{aligned}	  
\end{equation*}
Here, $D_Y$ ($D_X$) is the discriminator associated with the generator $G_{F}$ ($G_{R}$), and the scaling factor $\lambda_{cyc}=0.1$ is an empirically tuned hyper-parameter. We do not enforce additional global similarity loss-term because the $\mathcal{L}_{cyc}$ computes analogous reconstruction loss for each domain in $L_1$ space.

\begin{figure*}[t]
	\centering
	\begin{subfigure}{0.48\textwidth}
		\centering
		\includegraphics[width=\linewidth]{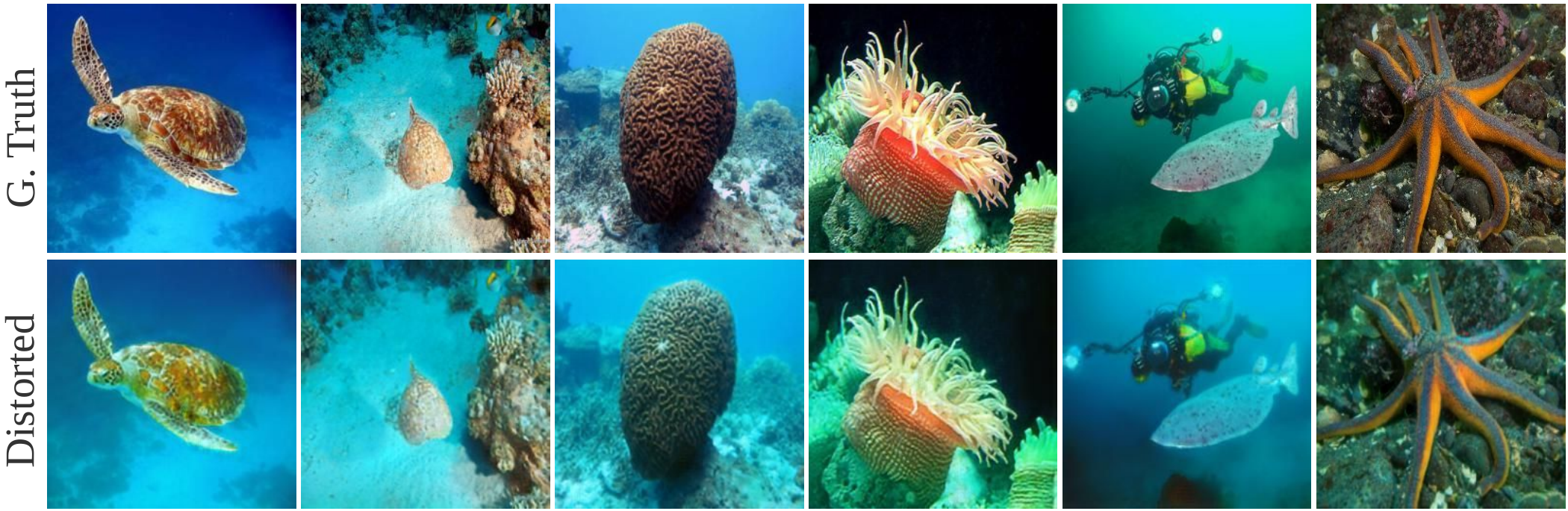} 
		\caption{Paired instances: ground truth images and their respective distorted pairs are shown on the top and bottom row, respectively.}
	\end{subfigure}~
	\begin{subfigure}{0.48\textwidth} 
		\centering
		\includegraphics[width=\linewidth]{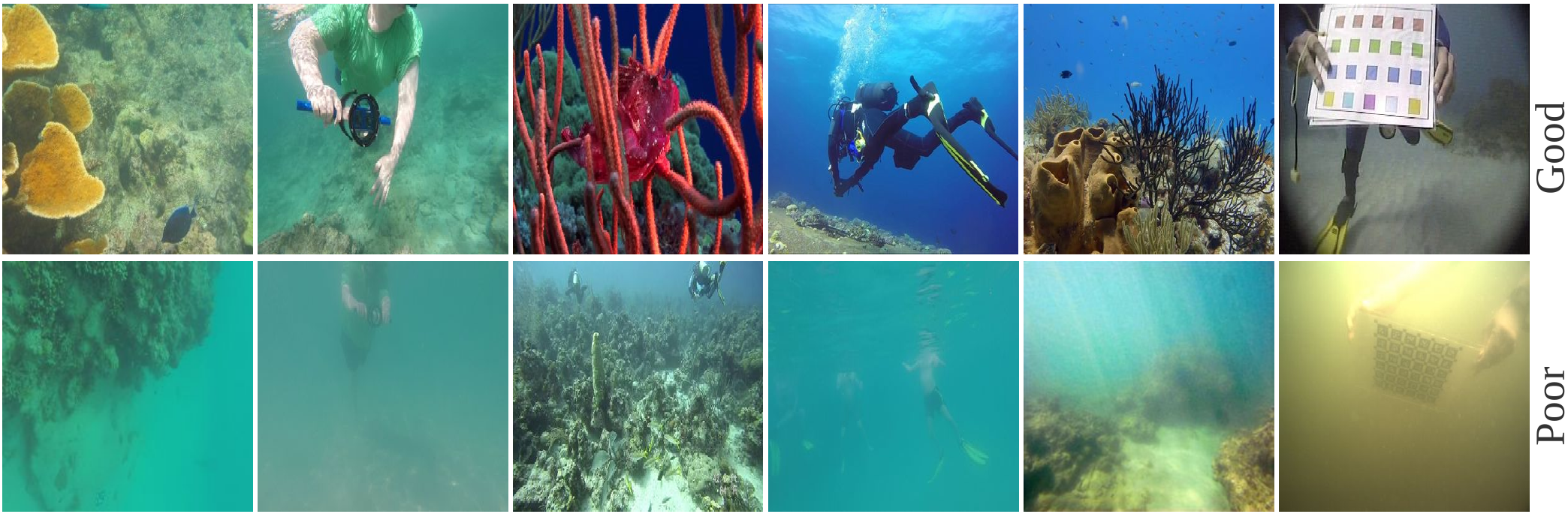} 
		\caption{Unpaired instances: good and poor quality images are shown on the top and bottom row (in no particular order), respectively.}
	\end{subfigure}
	
	%\vspace{1mm}
	\caption{A few sample images from the EUVP dataset are shown.}
	\label{fig:data}
\end{figure*}

\subsection{EUVP Dataset}
The EUVP dataset contains a large collection of paired and unpaired underwater images of poor and good perceptual quality. We used seven different cameras, which include multiple GoPros~\cite{gopro}, Aqua AUV's uEye cameras~\cite{dudek2007aqua}, low-light USB cameras~\cite{lowlight}, and Trident ROV's HD camera~\cite{trident}, to capture images for the dataset. The data was collected during oceanic explorations and human-robot cooperative experiments in different locations under various visibility conditions. Additionally, images extracted from a few publicly available YouTube\texttrademark videos are included in the dataset. 
The images are carefully selected to accommodate a wide range of natural variability (\eg, scenes, waterbody types, lighting conditions, etc.) in the data.

The unpaired data is prepared, \ie, good and poor quality images are separated based on visual inspection by six human participants. They inspected several image properties (\eg, color, contrast, and sharpness) and considered whether the scene is visually interpretable, \ie, foreground/objects are identifiable. Hence, the unpaired training endorses the modeling of human perceptual preferences of underwater image quality. On the other hand, the paired data is prepared by following a procedure suggested in~\cite{fabbri2018enhancing}. Specifically, a CycleGAN~\cite{zhu2017unpaired}-based model is trained on our unpaired data to learn the domain transformation between the good and poor quality images. Subsequently, the good quality images are distorted by the learned model to generate respective pairs; we also augment a set of underwater images from the ImageNet dataset~\cite{deng2009imagenet} and from Flickr\texttrademark.

There are over $12$K paired and $8$K unpaired instances in the EUVP dataset; a few samples are provided in Fig.~\ref{fig:data}. It is to be noted that our focus is to facilitate \textit{perceptual image enhancement} for boosting robotic scene understanding, not to model the underwater optical degradation process for \textit{image restoration}, which requires scene depth and waterbody properties.  
%The images are of various resolutions, \eg, $800 \times 600$, $640 \times 480$, $256 \times 256$, and $224 \times 224$. 

\section{Experimental Results}
%We now perform experimental evaluations of FUnIE-GAN using various qualitative analysis, standard quantitative metrics, and a user study.      
%\subsection{Training Processes}
We use TensorFlow libraries~\cite{abadi2016tensorflow} to implement the FUnIE-GAN model. It is trained separately on $11$K paired and $7.5$K unpaired instances; %that are randomly chosen from the EUVP dataset. 
the rest %, \ie, $1$K paired and $0.5$K unpaired images 
are used for respective validation and testing. Four NVIDIA\textsuperscript{TM} GeForce GTX 1080 graphics cards are used for training; both models are trained for $60$K-$70$K iterations with a batch-size of $8$. We now present the experimental evaluations based on a qualitative analysis, standard quantitative metrics, and a user study.

\subsection{Qualitative Evaluations}
\begin{figure}
	\centering
	\begin{subfigure}{0.49\textwidth}
		\centering
		\includegraphics[width=0.9\linewidth]{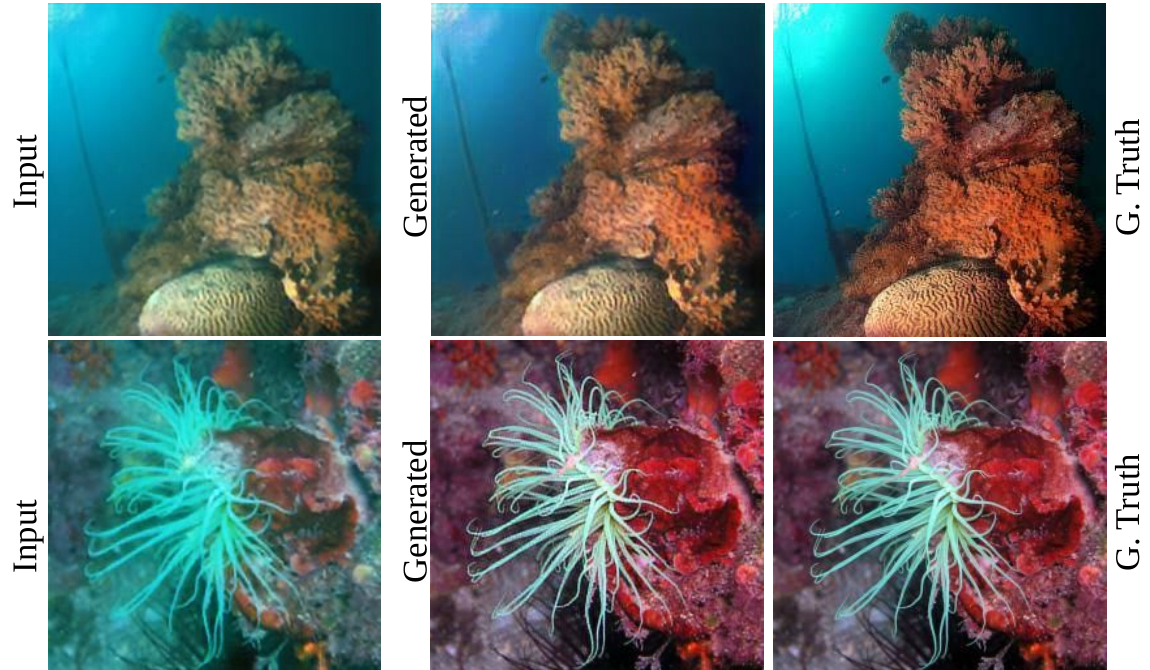}%
		\caption{True color and sharpness is restored in the enhanced image.}
		\label{fig:res2a}
	\end{subfigure}
	
	\vspace{1mm}
	\begin{subfigure}{0.49\textwidth}
		\centering
		\includegraphics[width=0.94\linewidth]{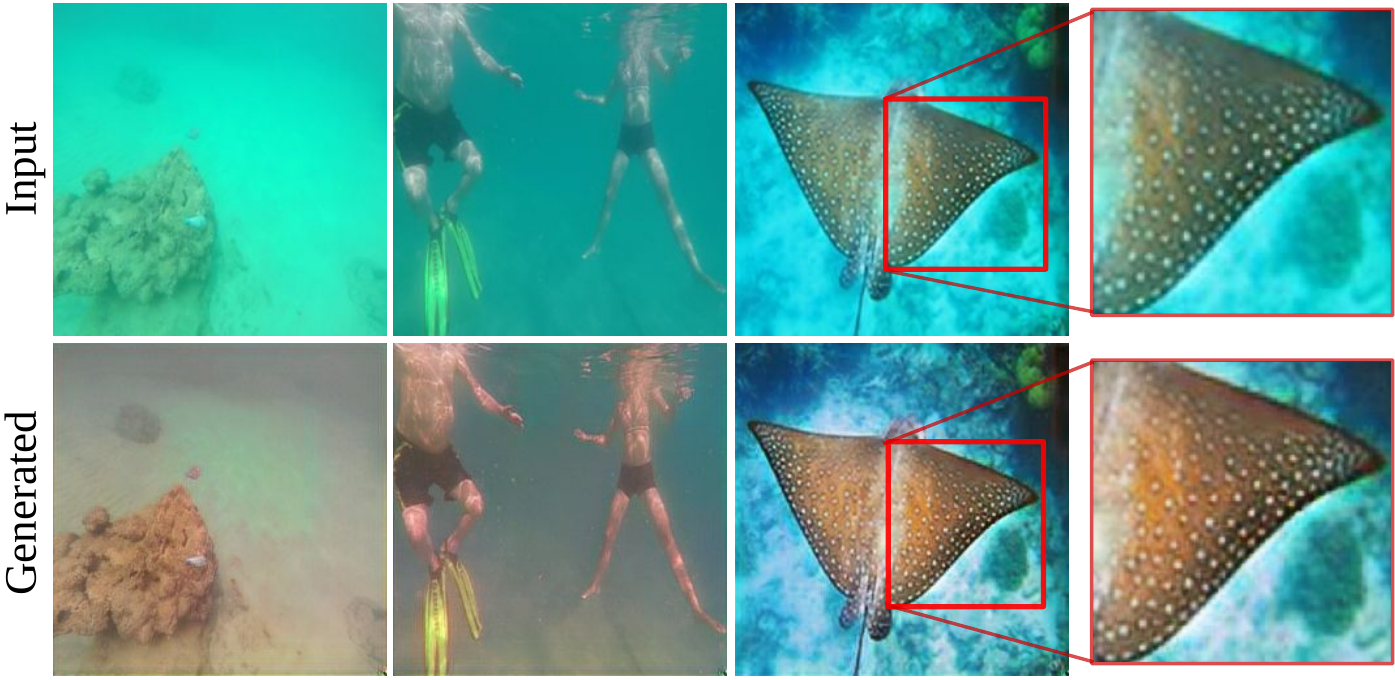}%
		\caption{The greenish hue is rectified and global contrast is enhanced.}
		\label{fig:res2b}
	\end{subfigure}
	
	\vspace{1mm}
	\begin{subfigure}{0.49\textwidth}
		\centering
		\includegraphics[width=0.96\linewidth]{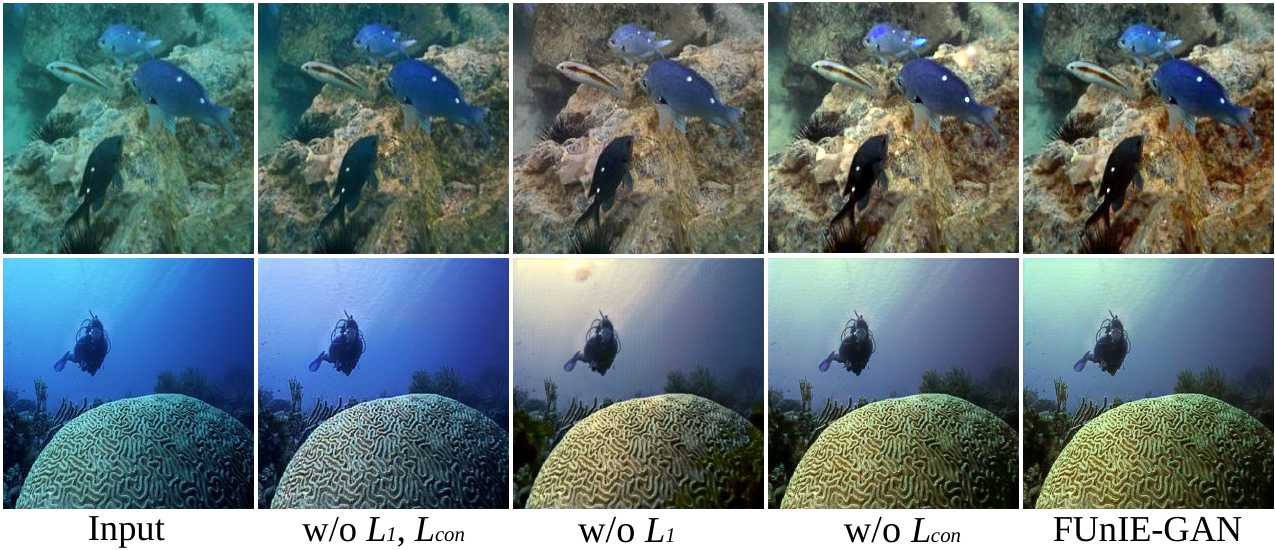}
		\caption{Ablation experiment: learning enhancement without (w/o) $\mathcal{L}_{1}$ and $\mathcal{L}_{con}$, w/o $\mathcal{L}_{1}$, and w/o $\mathcal{L}_{con}$ loss-terms in FUnIE-GAN.}
		\label{fig:res2c}
	\end{subfigure}
	
	\caption{Demonstration of improved image attributes by FUnIE-GAN in terms of color, sharpness, and contrast.}
	\vspace{-1mm}
	\label{fig:res1}
\end{figure}

We first qualitatively analyze the enhanced color and sharpness of the FUnIE-GAN-generated images compared to their respective ground truths. As Fig.~\ref{fig:res2a} shows, the true color, and sharpness is mostly recovered in the enhanced images. Additionally, as shown in Fig.~\ref{fig:res2b}, the greenish hue in underwater images are rectified and the global contrast is enhanced. These are the primary characteristics of an effective underwater image enhancer. 
We further demonstrate the contributions of each loss-terms of FUnIE-GAN: global similarity loss ($\mathcal{L}_{1}$), and image content loss ($\mathcal{L}_{con}$), for learning the enhancement. We observe that the $\mathcal{L}_{1}$ term helps to generate sharper images, while the $\mathcal{L}_{con}$ term contributes to furnishing finer texture details (see Fig.~\ref{fig:res2c}). Moreover, we found slightly better numeric stability for $\mathcal{L}_{con}$ with the {\tt block5\_conv2} layer of VGG-19 compared to its last feature extraction layer ({\tt block5\_conv4}). 

%\begin{figure}[h]
%    \centering
%    \includegraphics[width=0.95\linewidth]{figs/comp3.pdf}
%    \caption{Qualitative performance comparison of FUnIE-GAN and FUnIE-GAN-UP with physics-based methods: }
%    \label{fig:comp2}
%\end{figure}

\begin{figure*}[t]
	\centering
	\includegraphics[width=0.96\linewidth]{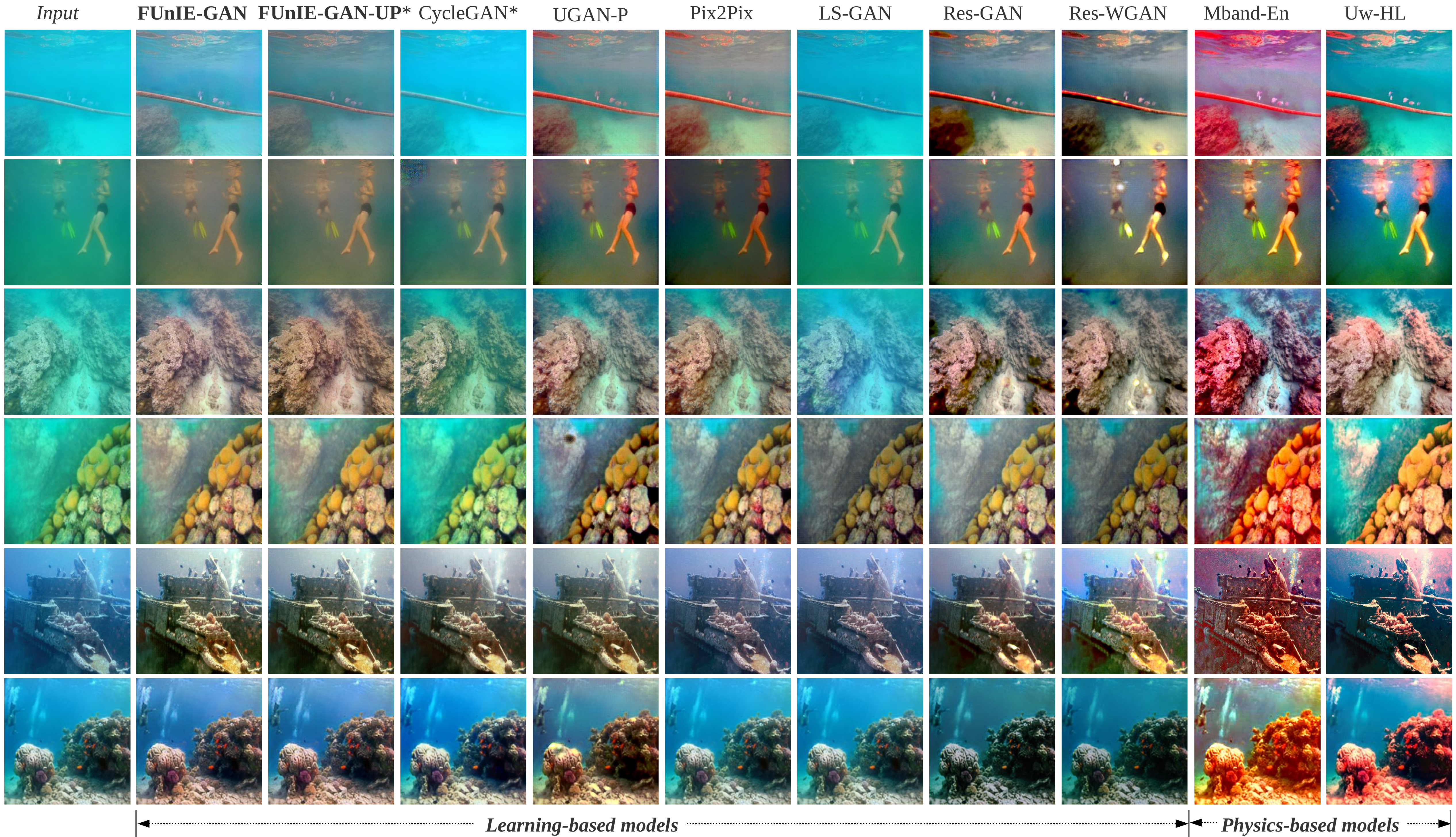}
	%\vspace{-1mm}
	\caption{Qualitative performance comparison of FUnIE-GAN and FUnIE-GAN-UP with learning-based methods: CycleGAN~\cite{zhu2017unpaired}, UGAN-P~\cite{fabbri2018enhancing}, Pix2Pix~\cite{isola2017image}, LS-GAN~\cite{mao2017least}, Res-GAN~\cite{li2017perceptual}, and Res-WGAN~\cite{arjovsky2017wasserstein}; the super-scripted asterisk ($\mathbf{\ast}$) denotes unpaired training. Two physics-based models: Mband-EN~\cite{cho2018model} and Uw-HL~\cite{berman2018underwater}, are also included in the comparison. (Best viewed at $400\%$ zoom)}
	\label{fig:comp}
\end{figure*}

Next, we conduct a qualitative comparison of perceptual image enhancement by FUnIE-GAN with several state-of-the-art models. We consider five learning-based models: (i) underwater GAN with gradient penalty (UGAN-P~\cite{fabbri2018enhancing}), (ii) Pix2Pix~\cite{isola2017image}, (iii) least-squared GAN (LS-GAN~\cite{mao2017least}), (iv) GAN with residual blocks~\cite{li2017perceptual} in the generator (Res-GAN), and (v) Wasserstein GAN~\cite{arjovsky2017wasserstein} with residual blocks in the generator (Res-WGAN). These models are implemented with $8$ encoder-decoder pairs (or $16$ residual blocks) in the generator network and $5$ convolutional layers in the discriminator. They are trained on the paired EUVP dataset using the same setup as the FUnIE-GAN. Additionally, we consider CycleGAN~\cite{zhu2017unpaired} as a baseline for comparing the performance of FUnIE-GAN with unpaired training (\ie, FUnIE-GAN-UP). We also include two physics-based models in the comparison: Multi-band fusion-based enhancement (Mbad-EN~\cite{cho2018model}), and haze-line-aware color restoration (Uw-HL~\cite{berman2018underwater}).   
A common test set with $1$K images (of $256\times256$ resolution) are used for the qualitative evaluation; it also includes $72$ images with known waterbody types~\cite{berman2018underwater}. A few sample comparisons are illustrated in Fig.~\ref{fig:comp}.

As demonstrated in Fig.~\ref{fig:comp}, Res-GAN, Res-WGAN, and Mbad-EN often suffer from over-saturation, while LS-GAN generally fails to rectify the greenish hue in images. UGAN-P, Pix2Pix, and Uw-HL perform reasonably well and their enhanced images are comparable to that of FUnIE-GAN; however, UGAN-P often over-saturates bright objects in the scene while Pix2Pix fails to enhance global brightness in some cases. 
On the other hand, we observe that achieving color consistency and hue rectification are relatively more challenging through unpaired learning. This is mostly because of the lack of reference color or texture information in the loss function. Nevertheless, FUnIE-GAN-UP still outperforms CycleGAN in general.     
Overall, FUnIE-GAN performs as well and often better without using scene depth or prior waterbody information as the physics-based models, and despite having a much simpler network architecture compared to the existing learning-based models.

\begin{table}[b]
\centering
\caption{Quantitative comparison for average PSNR and SSIM values on $1$K paired test images of EUVP dataset.}
\scriptsize
\vspace{-1mm}
\begin{tabular}{l||c|c}
  \hline
  \textbf{Model} & $PSNR\big(G(\mathbf{x}),\mathbf{y}\big)$ & $SSIM\big(G(\mathbf{x}),\mathbf{y}\big)$ \\  
   & Input: $17.27 \pm 2.88$ & Input: $0.62 \pm 0.075$  \\ \hline \hline
  Uw-HL & $18.85 \pm 1.76$  &  $0.7722 \pm 0.066$  \\ \hline 
  Mband-EN & $12.11 \pm 2.55$  &  $0.4565 \pm 0.097$ \\ \hline
  Res-WGAN & $16.46 \pm 1.80$ & $0.5762 \pm 0.014$ \\ \hline
  Res-GAN & $14.75 \pm 2.22$ & $0.4685 \pm 0.122$ \\ \hline
  LS-GAN & $17.83 \pm 2.88$ & $0.6725 \pm 0.062$ \\ \hline
  Pix2Pix & $20.27 \pm 2.66$ & $0.7081 \pm 0.069$ \\ \hline
  UGAN-P & $19.59 \pm 2.54$ & $0.6685 \pm 0.075$ \\ \hline
  CycleGAN & $17.14 \pm 2.65$ & $0.6400 \pm 0.080$  \\ \hline
  \textbf{FUnIE-GAN-UP} & $21.36 \pm 2.17$ & $0.8164 \pm 0.046$ \\ \hline
  \textbf{FUnIE-GAN} & $21.92 \pm 1.07$ & $0.8876 \pm 0.068$ \\ \hline
\end{tabular}
\label{tab:psnr_ssim}
%\vspace{-3mm}
\end{table}

\subsection{Quantitative Evaluation}
We consider two standard metrics~\cite{ignatov2017dslr,chen2018deep,hore2010image} named Peak Signal-to-Noise Ratio (PSNR) and Structural Similarity (SSIM) in order to quantitatively compare FUnIE-GAN-enhanced images with their respective ground truths. The PSNR approximates the reconstruction quality of a generated image $\mathbf{x}$ compared to its ground truth $\mathbf{y}$ based on their Mean Squared Error (MSE) as follows:
\begin{equation}
\begin{aligned}
\footnotesize
    %MSE(\mathbf{x},\mathbf{y})&=\frac{1}{mn} \sum_{i=1}^{m} \sum_{j=1}^{n} |\mathbf{x}_{i,j}-\mathbf{y}_{i,j}|^2 \\
    PSNR(\mathbf{x}, \mathbf{y}) &= 10 \log_{10} \big[255^2/MSE(\mathbf{x},\mathbf{y})\big] 
\end{aligned}
\label{eq:psnr}
\end{equation}

On the other hand, the SSIM~\cite{wang2004image} compares the image patches based on three properties: luminance, contrast, and structure. It is defined as follows:  
\begin{equation}
\footnotesize
        SSIM(\mathbf{x},\mathbf{y}) =  \Big(\frac{2 \mathbf{\mu}_\mathbf{x} \mathbf{\mu}_\mathbf{y} + c_1}{\mathbf{\mu}_\mathbf{x}^2+\mathbf{\mu}_\mathbf{y}^2 + c_1}\Big) \Big(\frac{2 \mathbf{\sigma}_{\mathbf{xy}} + c_2}{\mathbf{\sigma}_\mathbf{x}^2+\mathbf{\sigma}_\mathbf{y}^2 + c_2}\Big) 
\label{eq:ssm}
\end{equation}

In Eq.~\ref{eq:ssm}, $\mathbf{\mu}_\mathbf{x}$ ($\mathbf{\mu}_\mathbf{y}$) denotes the mean, and $\mathbf{\sigma}_\mathbf{x}^2$ ($\mathbf{\sigma}_\mathbf{y}^2$) is the variance of $\mathbf{x}$ ($\mathbf{y}$); whereas $\mathbf{\sigma}_{\mathbf{xy}}$ denotes the cross-correlation between $\mathbf{x}$ and $\mathbf{y}$. Additionally, $c_1 = (255\times0.01)^2$ and $c_2 = (255\times0.03)^2$ are constants that ensure numeric stability.

\begin{table}
\centering
\caption{Quantitative comparison for average UIQM values on $1$K paired and $2$K unpaired  test images of EUVP dataset.}
\scriptsize
\begin{tabular}{l||c|c}
  \hline
   & \textbf{Paired data} & \textbf{Unpaired data} \\ 
  \textbf{Model} & Input: $2.20 \pm 0.69$ & Input: $2.29 \pm 0.62$  \\
  & G. Truth: $2.91 \pm 0.65$ & G. Truth: N/A  \\ \hline \hline 
  Uw-HL & $2.62 \pm 0.35$ & $2.75 \pm 0.32$ \\ \hline 
  Mband-EN & $2.28 \pm 0.87$  & $2.34 \pm 0.45$ \\ \hline
  Res-WGAN & $2.55 \pm 0.64$ & $2.46 \pm 0.67$ \\ \hline
  Res-GAN & $2.62 \pm 0.89$ & $2.28 \pm 0.34$ \\ \hline
  LS-GAN & $2.37 \pm 0.78$ & $2.59 \pm 0.52$ \\ \hline
  Pix2Pix & $2.65 \pm 0.55$ & $2.76 \pm 0.39$ \\ \hline
  UGAN-P & $2.72 \pm 0.75$ & $2.77 \pm 0.34$ \\ \hline
  CycleGAN & $2.44 \pm 0.71$ & $2.62 \pm 0.67$ \\ \hline
  \textbf{FUnIE-GAN-UP} & $2.56 \pm 0.63$ & $2.81 \pm 0.65$ \\ \hline
  \textbf{FUnIE-GAN} &  $2.78 \pm 0.43$ & $2.98 \pm 0.51$ \\ \hline
\end{tabular}
\label{tab:uiqm}
\vspace{-3mm}
\end{table}

%We use a set of $1$K paired test images $(\mathbf{x} \in X, \mathbf{y} \in Y)$ in our evaluation. At first, we use FUnIE-GAN to generate enhanced images $G(\mathbf{x})$ for each $\mathbf{x}$ and then compute $PSNR\big(G(\mathbf{x}),\mathbf{y}\big)$ and $SSIM\big(G(\mathbf{x}),\mathbf{y}\big)$ using Eq.~\ref{eq:psnr} and~\ref{eq:ssm}, respectively. 
In Table~\ref{tab:psnr_ssim}, we provide the averaged PSNR and SSIM values over $1$K test images for FUnIE-GAN and compare the results with the same models used in the qualitative evaluation. The results indicate that FUnIE-GAN performs best on both PSNR and SSIM metrics. 
We conduct a similar analysis for Underwater Image Quality Measure (UIQM)~\cite{panetta2016human,liu2019real}, which %is a linear combination of three metrics: 
quantifies underwater image colorfulness, sharpness, and contrast. 
%We follow the relevant procedures for computing UIQM as described in~\cite{panetta2016human,liu2019real}. 
We present the results in Table~\ref{tab:uiqm}, which indicates that although FUnIE-GAN-UP performs better than CycleGAN, its UIQM values on the the paired dataset are relatively poor. Interestingly, the models trained on paired data, particularly FUnIE-GAN, UGAN-P, and Pix2Pix, produce better results. %perform well on both paired and unpaired test images. 
We postulate that the global similarity loss in FUnIE-GAN and Pix2Pix, or the gradient-penalty term in UGAN-P contribute to this enhancement, as they all add $L_{1}$ terms in the adversarial objective. Our ablation experiments of FUnIE-GAN (see Fig.~\ref{fig:res2c}) reveal that the $\mathcal{L}_{1}$ loss-term contributes to $4.58\%$ improvements in UIQM, while $\mathcal{L}_{con}$ contributes $1.07\%$. Moreover, without both $\mathcal{L}_{1}$ and $\mathcal{L}_{con}$ loss-terms, the average UIQM values drop by $17.6\%$; we observe similar statistics for PSNR and SSIM as well.

%The UIQM is expressed as follows:  
%\begin{equation}
%\begin{aligned}
%\footnotesize
%    UIQM(\mathbf{x}) = c_1 &\cdot UICM(\mathbf{x}) + \text{ } c_2 \cdot  UISM(\mathbf{x})  \\ 
%    &+ \text{ } c_3 \cdot UIConM(\mathbf{x})
%\end{aligned}
%\label{eq:uiqm}
%\end{equation}

%In Eq.~\ref{eq:uiqm}, the constant values are $c_1=0.0282$, $c_2=0.2953$, and $c_3=3.5753$; We follow the standard definition of Eq.~\ref{eq:uiqm} and relevant procedures for computing UICM, UISM, and UIConM that are described in~\cite{panetta2016human,liu2019real}. We present the results in Table~\ref{tab:uiqm}, which indicates that although FUnIE-GAN-UP performs better than CycleGAN, its UIQM values on the the paired dataset is relatively poor. Interestingly, the models trained on paired data, particularly FUnIE-GAN, UGAN-P, and Pix2Pix, perform well on both paired and unpaired test images. We postulate that the global similarity loss in FUnIE-GAN and Pix2Pix, or the gradient-penalty term in UGAN-P contribute to this enhancement, as they all add an $L_{1}$ term in the adversarial objective.               

%\begin{table}
%\centering
%\caption{Inference-time of FUnIE-GAN and UGAN-P on different hardware (in milliseconds).}
%\footnotesize
%\vspace{-1mm}
%\begin{tabular}{l||c|c|c}
%  \hline
%  \textbf{Model} & Titan Xp & Jetson TX2 & Robot CPU \\ \hline \hline
%  \textbf{FUnIE-GAN} &  $11.10$ ($90$ fps) &  $65.81$ ($15.2$ fps) & $126$ ($7.9$ fps) \\
%  \hline
%    UGAN-P &  $19.23$ ($52$ fps) &  $370.3$ ($2.7$ fps) & $-$ \\ \hline
%\end{tabular}
%\label{tab:time}
%\vspace{-3mm}
%\end{table}%%%%%%%%%%%%%%%%%%%%%%%%%%%%%%%%

\begin{table}[b]
\centering
\vspace{-2mm}
\caption{Rank-$n$ accuracy ($n=1,2,3$) for the top four models based on $312$ responses provided by $78$ individuals.}
\scriptsize
\begin{tabular}{l||c|c|c}
  \hline
  \textbf{Model} & Rank-1 ($\%$) & Rank-2 ($\%$) & Rank-3 ($\%$) \\ \hline %\hline
  \textbf{FUnIE-GAN} & $24.50$  & $68.50$   & $88.60$  \\ %\hline
  \textbf{FUnIE-GAN-UP} & $18.67$  & $48.25$   & $76.18$  \\ %\hline
   UGAN-P & $21.25$ & $65.75$  &  $80.50$ \\ %\hline
   Pix2Pix & $11.88$ & $45.15$  &  $72.45$ \\ \hline
\end{tabular}
\label{tab:study}
%\vspace{-2mm}
\end{table}%%%%%%%%%%%%%%%%%%%%%%%%%%%%%%%%

\subsection{User Study}
We also conduct a user study to add human preferences to our qualitative performance analysis. %As Fig.~\ref{fig:study} illustrates, 
The participants are shown different sets of $9$ images (one for each learning-based models) and asked to rank top $3$ best quality images. A total of $78$ individuals participated in the study and a total of $312$ responses are recorded. Table~\ref{tab:study} compares the average rank-1, rank-2, and rank-3 accuracy of the top $4$ categories. The average rank-3 accuracy of the original images is recorded to be $6.67$, which suggests that the users clearly preferred enhanced images over the original ones. Moreover, the results indicate that the users prefer the images enhanced by FUnIE-GAN, UGAN-P, and Pix2Pix compared to the other models; these statistics are consistent with our qualitative and quantitative analysis.

\begin{figure}[t]
    \centering
    \begin{subfigure}{0.49\textwidth}
    \centering
        \includegraphics[width=0.98\linewidth]{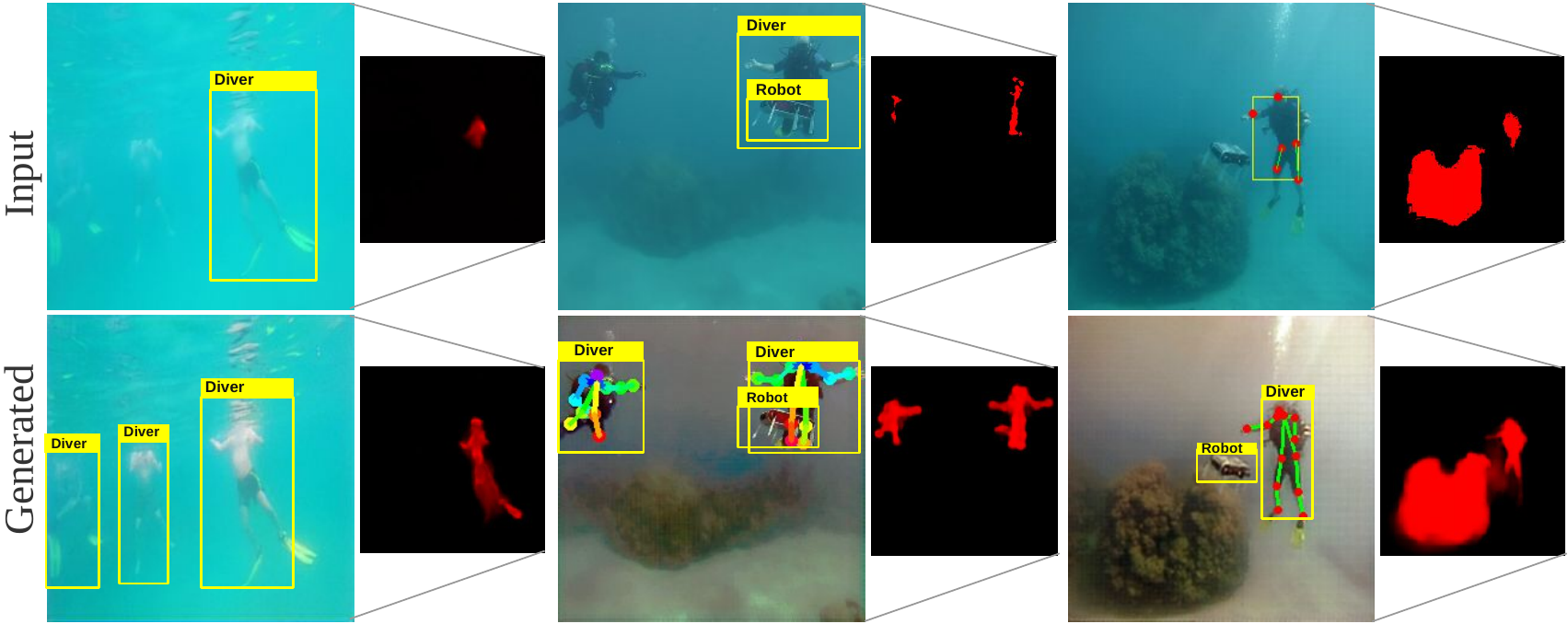} 
        \vspace{-1mm}
        \caption{A few snapshots showing qualitative improvement on FUnIE-GAN-generated images; a detailed demonstration can be found at: {\tt \url{https://youtu.be/1ewcXQ-jgB4}}.}
    \label{fig:per_a}
    \end{subfigure}

   \vspace{2mm}
    \begin{subfigure}{0.49\textwidth}
       \centering
        \includegraphics[width=0.98\linewidth]{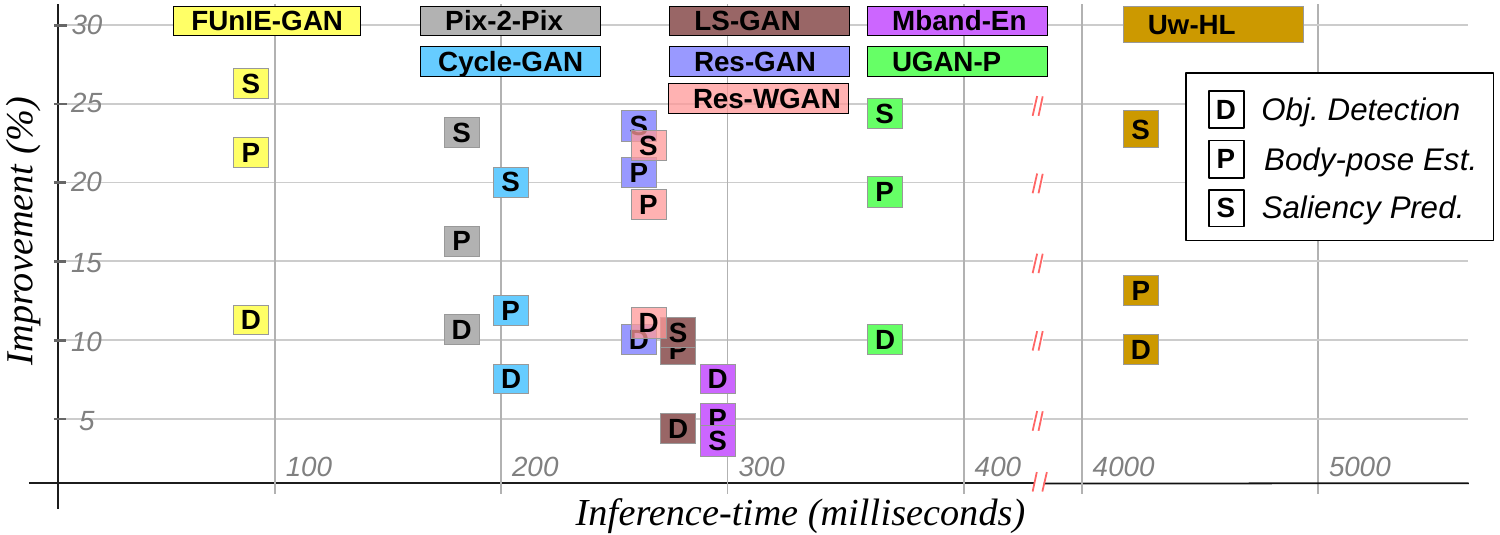}
    \vspace{-1mm}
    \caption{Improvement versus inference-time comparison with the state-of-the-art models; FUnIE-GAN offers over 10 FPS speed (on common platform: Intel\texttrademark Core-i5 3.6GHz CPU); note that the run-times are evaluated on $256\times256$ image patches for all the models.}
    \label{fig:per_b}
    \end{subfigure}
    \caption{Improved performance for object detection, saliency prediction, and human body-pose estimation on enhanced images.}
    %\vspace{-2mm}
    \label{fig:per}
\end{figure}

\subsection{Improved Visual Perception}
As demonstrated in Fig.~\ref{fig:per_a}, we conduct further experiments to quantitatively interpret the effectiveness of FUnIE-GAN-enhanced images for underwater visual perception over a variety of test cases. We analyze the performance of standard deep visual models for underwater object detection~\cite{islam2018towards}, 2D human body-pose estimation~\cite{cao2017realtime}, and visual attention-based saliency prediction~\cite{wang2018salient}; although results vary depending on the image qualities of a particular test set, on an average, we observe $11$-$14\%$, $22$-$28\%$, and $26$-$28\%$ improvements, respectively. We also evaluate other state-of-the-art models on the same test sets; as Fig.~\ref{fig:per_b} suggests, images enhanced by UGAN-P, Res-GAN, Res-WGAN, Uw-HL, and Pix2Pix also achieve considerable performance improvements. However, these models offer significantly slower inference-rates than FUnIE-GAN, most of which are not suitable for real-time deployment in robotic platforms. 

FUnIE-GAN's memory requirement is $17$ MB and it operates at a rate of $25.4$ FPS (frames per second) on a single-board computer (NVIDIA\texttrademark Jetson TX2), $148.5$ FPS on a graphics card (NVIDIA\texttrademark GTX 1080), and $7.9$ FPS on a robot CPU (Intel\texttrademark Core-i3 6100U). 
These computational aspects are ideal for it to be used as an image processing pipeline by visually-guided underwater robots in real-time applications. 

%\begin{figure}[b]
%    \centering
%        \includegraphics[width=0.8\linewidth]{figs/study.pdf}
%    %\vspace{-1mm}
%    \caption{A snapshot of the user interface used in our study.}
%    \label{fig:study}
%\end{figure}

\subsection{Limitations and Failure Cases}\label{limit}
\begin{figure}

    \centering
        \includegraphics[width=0.98\linewidth]{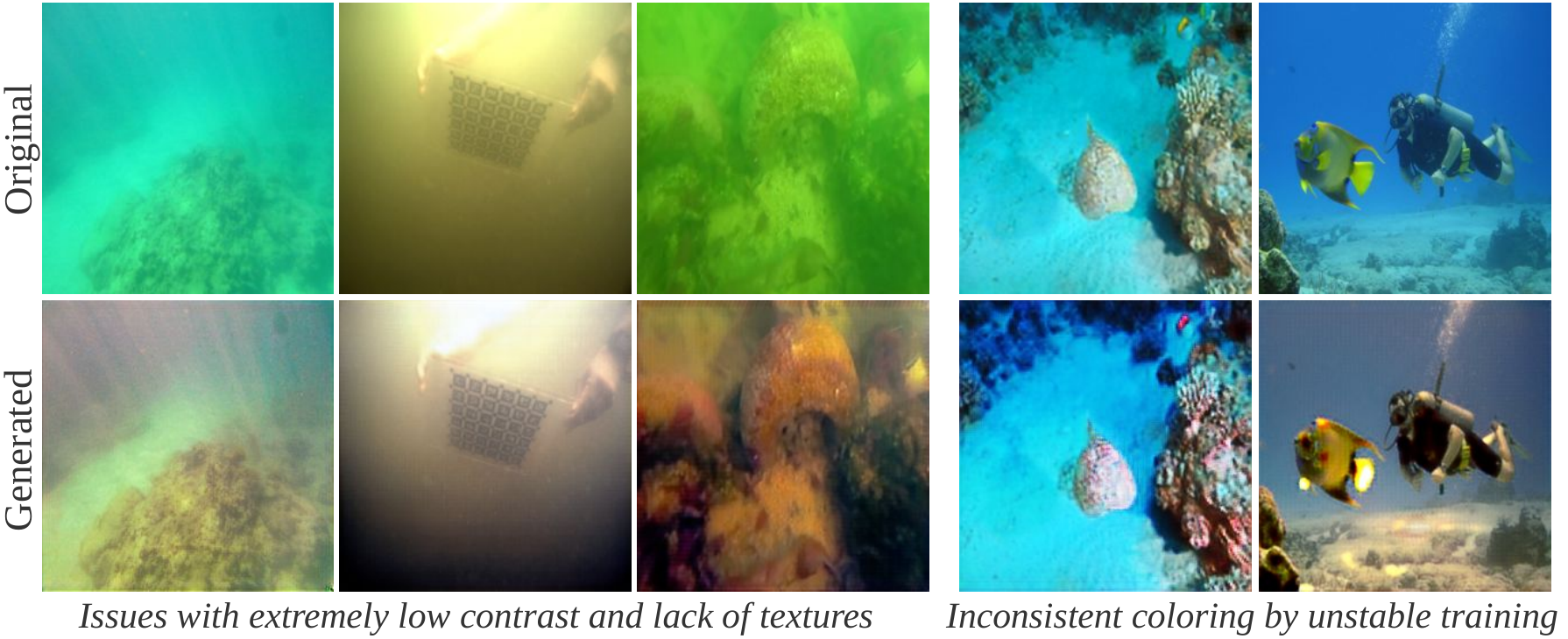}
    \caption{Extremely low-contrast and texture-less images are generally challenging for FUnIE-GAN, whereas FUnIE-GAN-UP often suffers from inconsistent coloring due to training instability.}
    \label{fig:bad}
    \vspace{-2mm}
\end{figure}

We observe a couple of challenging cases for FUnIE-GAN, which are depicted by a few examples in Fig.~\ref{fig:bad}. First, FUnIE-GAN is not very effective for enhancing severely degraded and texture-less images. The generated images in such cases are often over-saturated by noise amplification. Although the hue rectification is generally correct, the color and texture recovery remains poor. Secondly, FUnIE-GAN-UP is prone to training instability. Our investigations suggest that the discriminator often becomes too good too early, causing a \textit{diminishing gradient} effect that halts the generator's learning. As shown in Fig.~\ref{fig:bad} (right), the generated images in such cases lack color consistency and accurate texture details. This is a fairly common issue in unpaired training of GANs~\cite{chen2018deep,ignatov2017dslr,johnson2016perceptual}, and requires meticulous hyper-parameter tuning.     

FUnIE-GAN balances a trade-off between robustness and efficiency which limits its performance to a certain degree. More powerful deep models (\ie, denser architectures with more parameters) can be adopted for non-real-time applications; moreover, the input/output layers can be modified with additional bottleneck layers for learning enhancement at higher resolution than $256\times256$. On the other hand, FUnIE-GAN does not guarantee the recovery of true pixel intensities as it is designed for perceptual image quality enhancement. If scene depth and optical waterbody properties are available, underwater light propagation and image formation characteristics~\cite{akkaynak2018revised,berman2018underwater,bryson2016true} can be incorporated into the optimization for more accurate image restoration.

\section{Conclusion}
We present a simple yet efficient conditional GAN-based model for underwater image enhancement. The proposed model formulates a perceptual loss function by evaluating image quality based on its global color, content, local texture, and style information. We also present a large-scale dataset containing a paired and an unpaired collection of underwater images for supervised training. We perform extensive qualitative and quantitative evaluations, and conduct a user study which show that the proposed model performs as well and often better compared to the state-of-the-art models, in addition to ensuring much faster inference time. Moreover, we demonstrate its effectiveness in improving underwater object detection, saliency prediction, and human body-pose estimation performances. %These results suggest that the proposed model can be used to improve real-time perception performances of visually-guided underwater robots. 
In the future, we plan to investigate its feasibility in other underwater human-robot cooperative applications, marine trash identification, etc. We seek to improve its color consistency and stability for unpaired training as well. 
%%%%%%%%%%%%%%%%%%%%%%%%%%%%%%

%%%%%%%%%%%%%%%%%%%%%%%%%%%%%%%%%%%%%%%%%%%%%
{\small
\bibliographystyle{ieee}
\bibliography{Refs}
}
%%%%%%%%%%%%%%%%%%%%%%%%%%%%%%%%%%%%%%%%%%%%%

\end{document}